\documentclass{article}
\usepackage{ijcai18}

\usepackage{times}
\usepackage{soul}
\usepackage{url}
\usepackage[hidelinks]{hyperref}
\usepackage[utf8]{inputenc}
\usepackage[small]{caption}
\usepackage{amsmath}
\usepackage{graphicx}
\usepackage{tabularx}

\title{Cross-Lingual Task-Specific Representation Learning for Text Classification in Resource Poor Languages \thanks{This work was presented at 1st Workshop on Humanizing AI (HAI) at IJCAI'18 in Stockholm, Sweden.}}

\author{
Nurendra Choudhary$^1$,
Rajat Singh$^1$,
Manish Shrivastava$^1$
\\ 
$^1$ Language Technologies Research Center, Kohli Center on Intelligent Systems,\\
International Institute of Information Technology, Hyderabad, India.\\
\{nurendra.choudhary,rajat.singh\}@research.iiit.ac.in,
m.shrivastava@iiit.ac.in}

\begin{document}

\maketitle

\begin{abstract}
Neural network models have shown promising results for text classification. However, these solutions are limited by their dependence on the availability of annotated data. 

The prospect of leveraging resource-rich languages to enhance the text classification of resource-poor languages is fascinating. The performance on resource-poor languages can significantly improve if the resource availability constraints can be offset. To this end, we present a twin Bidirectional Long Short Term Memory (Bi-LSTM) network with shared parameters consolidated by a contrastive loss function (based on a similarity metric). The model learns the representation of resource-poor and resource-rich sentences in a common space by using the similarity between their assigned annotation tags. Hence, the model projects sentences with similar tags closer and those with different tags farther from each other. We evaluated our model on the classification tasks of sentiment analysis and emoji prediction for resource-poor languages - Hindi and Telugu and resource-rich languages - English and Spanish. Our model significantly outperforms the state-of-the-art approaches in both the tasks across all metrics.
\end{abstract}

\section{Introduction}
\label{sec:introduction}
Text classification is among the primary challenges in natural language processing. Several approaches, ranging from rule-based systems to machine learning algorithms, have been studied to tackle the problem. Deep learning approaches, especially, show exceptional effectiveness in the task. In this paper, we propose a unified architecture to harness resource-rich languages for the task of representation learning of a resource-poor language for a given task. The proposed architecture consists of two Bi-LSTM networks with mutually shared parameters consolidated by a contrastive loss function. The adopted energy function suits discriminative training for energy based models \cite{lecun2005loss}.

The model initiates with a simple one-hot representation based on character trigrams in the sentences. The shared parameters of the siamese network in conjunction with the similarity metric (adopted for the loss function) learn representations of the sentences in accordance to the task. The contrastive loss function projects sentences into the problem space such that sentences with similar tags are closer and different tags are farther from each other. 

The input sentences are language-agnostic as the representations only depend on the similarity between their tags. But we require that the tags of both the sentences should represent the same concept. This feature helps us in improving the performance of our model on resource-poor languages by leveraging the abundant resources for the same task available in other languages.
\section{Related Work}
\label{sec:related work}
Distributional semantic vectors \cite{mikolov2013distributed} partially capture the semantics of a sentence but ignore syntactic information and diverse word senses. \cite{mukku2016enhanced} utilizes a combination of semantic vectors and morphology analyzer to solve multi-class sentiment analysis in Telugu language.

Matrix Vector Recursive Neural Networks (MV-RNN) \cite{socher2012semantic} provides a solution that considers both individual meaning of a word and its semantic relation with other words in the sentence. The model assigns a vector and a matrix to each word which represent its semantic value and relation with other words respectively.
\begin{figure*}
\centering
\includegraphics[width=0.8\linewidth]{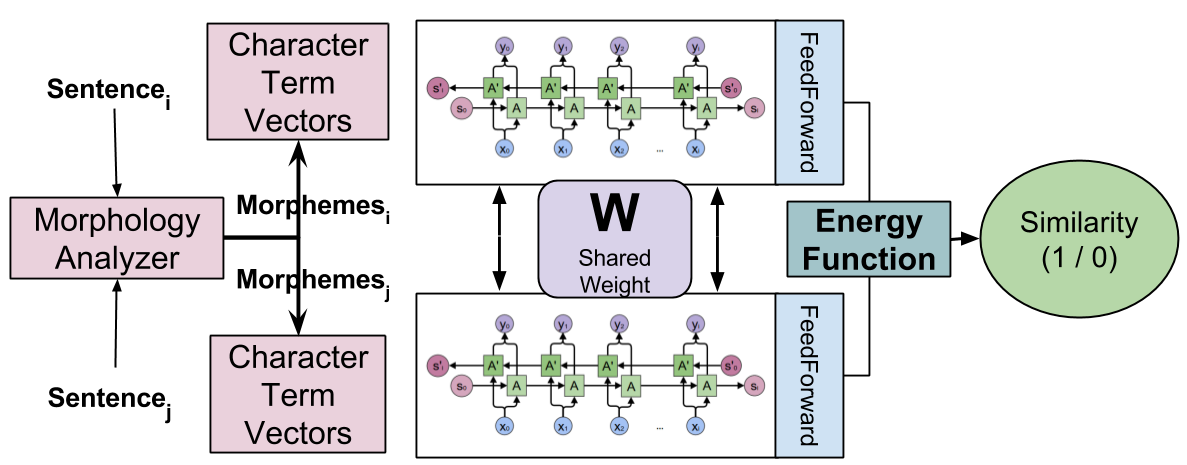}
\captionof{figure}{Architecture of the Model}
\label{fig:cesna}
\end{figure*}
Another line of research \cite{joshi2010fall,balamurali2012cross} benefits from the defined grammatical rules and vocabulary of Hindi language. Although, these solutions are highly accurate, they are susceptible to incorrect spellings, rare words or any exceptional structures frequently observed in informal social media data such as reviews and tweets.

For multilingual emoji prediction, Bi-LSTM model \cite{barbieri2017emojis} utilizes recurrent neural networks to capture the character and word sequences of the sentence. The model presents an effective approach to capture the sequence and content of the sentence. Also, it captures close proximity semantic relations in the sentence. However, training the model requires immense data. Thus, the system fails in case of inadequate data.

For informal data, methods that require immutable words are ineffective. Given their proven effectiveness \cite{dhingra2016tweet2vec,chung2014empirical,vinyals2015grammar,bahdanau2014neural,dyer2015transition,wang2016learning}, we adopt Bi-LSTMs based on sequences of character trigrams. This approach solves the problems of incorrect spellings, rare words and agglutination (in case of Hindi and Telugu).

Although, Bi-LSTMs project sentences into the problem space, we also need the sentences with similar properties closer and those with different properties farther in the problem space. To solve this problem, we utilize siamese networks. 

Siamese networks are capable of learning a similarity from given data without requiring specific information about the classes. \cite{bromley1994signature} introduced siamese networks for the task of signature verification. Recently, \cite{das2016together} utilized siamese networks to solve community question-answering. 

\section{Architecture of the Model}
\label{sec:architecture}
As depicted in figure \ref{fig:cesna}, the architecture consists of a Bi-LSTM pair connected to a dense feed-forward layer at the top. The Bi-LSTMs capture the sequence and content of the character trigrams in the sentence and project them into the problem space. The contrastive loss function combines the similarity and the sample's label. Back-propagation through time computes the loss function's gradient with respect to the weights and biases shared by the sub-networks. The weights are then updated to rectify the error in the epoch.
\subsection{Primary Representations}
Informal data consists of spelling errors, rare words and variations of the same word. The style of writing a word may also convey a feature (e.g; ``Hiiii" conveys a positive sentiment whereas ``Hi" is a neutral sentiment). Hence, we consider character trigrams to embed the sentence instead of words. This approach handles spelling errors and rare words. Character trigrams take the information of all the word's inflections, thus, eliminating the problem of agglutination. Also, this method captures the sentiment of different writing styles as information is attained on a character-level. To further address the problem of agglutination in morphologically rich languages, we add a morphology analyzer that segments the words into its constituent morphemes. The approach represents a sentence using a one-hot vector with number of dimensions equal to the number of unique character trigrams in the training dataset.

We input character-based term vectors of both the languages' sentences and a label to the twin networks of our model. The label indicates whether the samples are nearer or farther to each other in the problem space. For positive samples (nearer in the problem space), we feed the twin networks with term vectors of sentences with the same tag. For negative samples (far away from each other in the problem space), we feed the twin networks with term vectors of sentences with different tags.
\begin{figure*}
\centering
\includegraphics[width=0.6\linewidth]{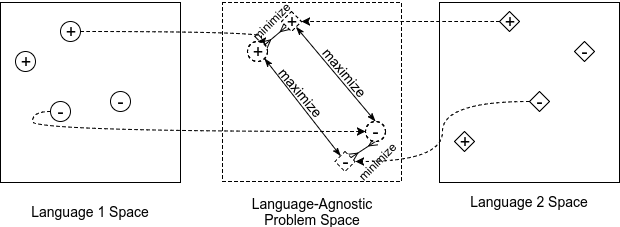}
\captionof{figure}{Projection of sentences in both the languages into a common problem space. Distance between similar samples is minimized and dissimilar samples in maximized.}
\label{fig:projection}
\end{figure*}
\begin{table*}[t]
\centering
\begin{tabular}{|m{0.5cm}|m{0.54cm}|m{0.54cm}|m{0.54cm}|m{0.35cm}|m{0.35cm}|m{0.5cm}|m{0.35cm}|m{0.35cm}|m{0.35cm}|m{0.35cm}|m{0.35cm}|m{0.5cm}|m{0.35cm}|m{0.35cm}|m{0.35cm}|m{0.35cm}|m{0.35cm}|m{0.35cm}|}
\hline
\textbf{Lan}&
\includegraphics[height=1.1em]{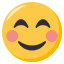}&  \includegraphics[height=1.1em]{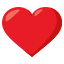}&    \includegraphics[height=1.1em]{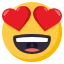}&\includegraphics[height=1.1em]{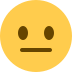}&
\includegraphics[height=1.1em]{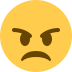}&
\includegraphics[height=1.1em]{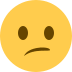}&
 \includegraphics[height=1.1em]{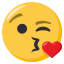}&
\includegraphics[height=1.1em]{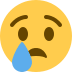}&
\includegraphics[height=1.1em]{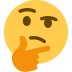}&
\includegraphics[height=1.1em]{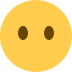}&\includegraphics[height=1.1em]{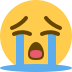}&\includegraphics[height=1.1em]{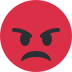}&
  \includegraphics[height=1.1em]{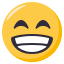}&
 \includegraphics[height=1.1em]{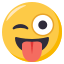}&
 \includegraphics[height=1.1em]{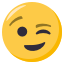}&
 \includegraphics[height=1.1em]{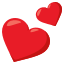}&\includegraphics[height=1.1em]{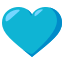}&\includegraphics[height=1.1em]{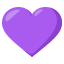}\\
 \hline
 Eng&17.1&15.3&10.0&5.7&4.9&4.7&3.8&3.7&3.6&3.5&3.3&3.2&3.1&3.0&2.8&2.7&2.7&2.6\\
 Spa&9.7&10.8&13.1&3.1&2.7&6.5&6.3&6.8&6.0&3.4&6.1&4.0&4.6&5.4&3.4&2.8&3.4&2.0\\
 Hin&9.7&6.8&7.5&3.9&2.8&9.5&4.9&1.5&6.7&2.9&4.3&9.4&8.6&7.7&7.2&1.9&2.3&2.6\\
 Tel&22.7&5.7&16.6&1.4&0.3&10.8&3.9&0.3&3.0&1.6&0.9&13.4&5.1&6.7&4.2&2.3&0.9&0.5\\
 \hline
\end{tabular}
\caption[]{Distribution of emojis in languages' tweets. \footnotemark}
\label{tab:emoji dist}
\end{table*}
\subsection{Bi-directional LSTM Network}
We map each sentence-pair into [$l^1_{i},l^2_{i}$] such that $l^1_{i} \in {\rm I\!R}^{m}$ and $l^2_{i} \in {\rm I\!R}^{n}$, where $m$ and $n$ are the total number of character trigrams in both the languages respectively.

Bi-LSTM model encodes the sentence twice, one in the original order (forward) of the sentence and one in the reverse order (backward). Back-propagation through time \cite{boden2002guide} calculates the weights for both the orders independently. The algorithm works in the same way as general back-propagation, except in this case the back-propagation occurs over all the hidden states of the unfolded timesteps.

We, then, apply element-wise Rectified Linear Unit (ReLU) to the output encoding of the BiLSTM. ReLU is defined as: $f (x) = max (0,x)$. The choice of ReLU simplifies back-propagation, causes faster learning and avoids saturation.

The architecture's final dense feed forward layer converts the output of the ReLU layer into a fixed length vector $s \in {\rm I\!R}^{d}$. In our architecture, we empirically set the value of $d$ to 128. The overall model is formalized as:
\begin{equation}
s=max\{0,W[fw,bw]+b\}
\end{equation}
where $W$ is a learned parameter matrix (weights), $fw$ is the forward LSTM encoding of the sentence, $bw$ is
the backward LSTM encoding of the sentence, and
$b$ is a bias term.
\begin{table*}
\centering
\begin{tabular}{c|c|ccc}
Datasets&Sentence Length&Positive&Negative&Neutral\\\hline
English - Movie Reviews&429&38\%&24\%&38\%\\
English - Twitter&12&29\%&26\%&45\%\\
Spanish - Twitter&14&48\%&13\%&39\%\\
Hindi - Reviews&15&33\%&31\%&36\%\\
Telugu - News&13&27\%&27\%&46\%\\
\end{tabular}
\caption{Distribution of the datasets considered for the sentiment analysis task.}
\label{tab:sentiment dist}
\end{table*}
\subsection{Training and Testing Phases}
We train our model on the pairs of sentences in both the languages to capture the similarity in their classes. Our architecture differs from other deep learning counterparts due to its property of parameter sharing. Training the network with a shared set of parameters not only reduces the number of parameters (thus, save many computations) but also ensures that the sentences of both the languages project into the same space (shown in figure \ref{fig:projection}). We learn the network's shared parameters to minimize the distance between the sentences with the same classes and maximize the distance between the sentences with different classes.

Given an input ${l^1_{i},l^2_{i}}$ where $l^1_{i}$ and $l^2_{i}$ are sentences from both the languages respectively and a label $y_{i}\in \{ -1,1 \}$, the loss function is defined as:
\begin{equation}
loss(l^1_{i},l^2_{i})=
\begin{cases} 
1-cos(l^1_{i},l^2_{i}),&\text{$y=1$};\\
max(0,cos (l^1_{i},l^2_{i})-m),&\text{$y=-1$};
\end{cases}
\end{equation}
where $m$ is the margin that decides the distance by which dissimilar pairs should be moved away from each other. It generally varies between 0 to 1. The loss function is minimized such that pair of sentences with the label 1 (same class) are projected nearer to each other and pair of sentences with the label -1 (different class) are projected farther from each other in the problem space.The model is trained by minimizing the overall loss function in a batch. The objective is to minimize:
\begin{equation}
L(\Lambda) = \sum_{(l^1_{i},l^2_{i})\in C \cup C'} loss(l^1_{i},l^2_{i})
\end{equation}
where $C$ contains the batch of same sentiment sentence pairs and $C'$
contains the batch of different sentiment sentence pairs. Back-propagation through time (BPTT) updates the parameters shared by the Bi-LSTM sub-networks.

For testing, we randomly sample a certain number (100 in our case) of sentences for each class $R_{class}$ from the language corpus with higher amount of data. For every input, we then apply the trained model to get the similarity between the input and all corresponding $R_{class}$. Finally, we select the $R_{class}$ with the most matches with the input as the correct tag.

In case the correlated data available for both the languages are not annotated, we utilize one language's abundant resources to construct a state-of-the-art sentiment analysis model. The sentiment analysis model in conjunction with the correlation data obtained from our model aids the resource-poor language's prediction.
\footnotetext{The hearts in the table are of different colors. The most frequent one is \textit{red}, the second most frequent one is \textit{blue} and the last one is \textit{purple heart}}
\section{Experiments}
\label{sec:experiments}
This section studies and evaluates the performance of our model. We also compare our model with the previous approaches in the domain. To perform a proper comparative study and evaluation, we test our system on the specific applications of text classification - multilingual sentiment analysis and multilingual emoji prediction.
\begin{table}
\centering
\begin{tabular}{c|cccc}
\hline
\textbf{Dataset}&English&Spanish&Hindi&Telugu\\\hline
\textbf{\#} of tweets&500,000&100,000&15000&6000\\\hline
\end{tabular}
\caption{Datasets for emoji prediction task.}
\label{tab:emoji dataset}
\end{table}

\subsection{Datasets}
We consider varied datasets for both the problems of sentiment analysis and emoji prediction. We also differentiate between languages on the basis of resources. Hindi and Telugu have significantly less resources compared to English and Spanish. Hence, Hindi and Telugu are considered resource-poor languages here and English and Spanish are considered resource-rich.
The datasets considered for the task of sentiment analysis are given in Table \ref{tab:sa dataset}. The  datasets for sentiment analysis are annotated into 3 sentiment classes - positive, neutral and negative. The sentiment tags' distribution in these datasets is given in table \ref{tab:sentiment dist}.
\begin{table}
\centering
\begin{tabular}{c|c}
\textbf{Dataset}&\textbf{\# of Sentences}\\
\hline
English-Movie Reviews&5006 movie reviews\\
\cite{pang2005seeing}&\\\hline
English-Twitter&103035 tweets\\
\cite{mozetivc2016multilingual}&\\\hline
Spanish-Twitter&275589 tweets\\
\cite{mozetivc2016multilingual}&\\\hline
Hindi-Product Reviews&1004 product reviews\\
\cite{mogadala2012retrieval}&\\\hline
Telugu-News Corpus&1644 news lines\\
\cite{mukku2016enhanced}&\\
\end{tabular}
\caption{Datasets for sentiment analysis. }
\label{tab:sa dataset}
\end{table}
The twitter-emoji datasets considered for the task of emoji prediction are given in Table \ref{tab:emoji dataset}. Tweets from the ids given by \cite{barbieri2017emojis} are used to create English dataset. The dataset consists of the tweets with 18 most frequent emojis in English.
Each tweet considered in the datasets consists of a single or multiple instances of a single type of emoji. Table \ref{tab:emoji dist} demonstrates the distribution of the emojis in the datasets for emoji prediction task.
\subsection{Baselines}
\label{sec:baselines}
The approaches vary based on the language in consideration. Hence, we accordingly define the baselines below.
English, Japanese and Spanish enjoy the highest share of data on Twitter\footnote{The Many Tongues of Twitter - MIT Technology Review}. We consider English and Spanish because of their script and typological similarity (both are Subject-Verb-Object).\\
\textbf{Average Skip-gram Vectors (ASV):} We train a Word2Vec skip-gram model \cite{mikolov2013distributed} on a corpus of 65 million raw sentences in English and 20 million raw sentences in Spanish. Word2Vec provides a vector for each word. We average the words' vectors to get the sentence's vector. After obtaining each message's embedding, we train an L2-regularized logistic regression (with $\epsilon$ equal to 0.001).\\
\textbf{Matrix Vector Recursive Neural Network (MV-RNN):} The model \cite{socher2012semantic} assigns a vector and matrix to the nodes of a syntactic parsed tree. The vector represents the node's semantic value and matrix represents its relation with neighboring words. A recursive neural network model is trained using back-propagation through structure to define the nodes' contribution to the sentence's sentiment.\\
\textbf{Bidirectional LSTM (Bi-LSTM):} There are two approaches - word based and character based Bi-LSTM embeddings. We model the architecture as described in \cite{barbieri2017emojis}. The Bi-LSTMs capture the features of the sequence of characters/words into vectors. We, then, use these feature vectors for classification of the sentences.

Hindi and Telugu are the $3^{rd}$ and $17^{th}$ most spoken language in the world respectively. But they hold a relatively low share of data. This also further translates to a limited availability of annotated corpus for these languages. The baselines for these languages are:\\
\textbf{Domain Specific Classifier (Telugu) (DSC-T):} We train a Word2Vec model on a corpus of 700,000 Telugu sentences provided by Indian Languages Corpora Initiative. We train a Random Forest (given by \cite{mukku2016enhanced}) on the Telugu News dataset to construct our baseline for Telugu language.\\
\textbf{Multinomial Naive Bayes Model (Hindi) (MNB-H):} We train a multinomial naive bayes model (given by \cite{sarkar2015sentiment}) on the Hindi Review dataset to form our baseline for Hindi language.

The model of MV-RNN is defined for the problem of sentiment analysis, so we do not consider them for the task of emoji prediction. Similarly, Bi-LSTM model is only compared in the emoji prediction task and not sentiment analysis.
\begin{table}[t]
\centering
\begin{tabular}{c|cccc}
\textbf{Method}&\textbf{Accuracy}&\textbf{Precision}&\textbf{Recall}&\textbf{F-score}\\
\hline
ASV&52.59\%&0.49&0.52&0.50\\
MV-RNN&79.0\%&0.77&0.75&0.76\\
DSC-T (RF)&67.17\%&0.67&0.66&0.66\\
MNB-H&62.14\%&0.61&0.58&0.59\\\hline
SSC(Eng-Eng)&\textbf{82.25\%}&\textbf{0.83}&\textbf{0.80}&\textbf{0.81}\\
SSC(Spa-Spa)&81.5\%&0.83&0.80&0.81\\
SSC(Hin-Hin)&70.2\%&0.72&0.69&0.70\\
SSC(Tel-Tel)&69.2\%&0.70&0.69&0.69\\
SSC(Hin-Eng)&\textbf{80.5\%}&\textbf{0.82}&\textbf{0.79}&\textbf{0.80}\\
SSC(Tel-Eng)&\textbf{80.3\%}&\textbf{0.82}&\textbf{0.79}&\textbf{0.80}\\
\end{tabular}
\caption{Comparison between different language pairs of our model and previous methodologies for sentiment analysis experiment. Siamese Sentiment Classifier (SSC) refers to our model.}
\label{tab:sentiment analysis}
\end{table}
\begin{table*}[t]
\centering
\begin{tabular}{c|ccc|ccc|ccc}
&&\textbf{5}&&&\textbf{10}&&&\textbf{18}&\\
\textbf{Method}&\textbf{P}&\textbf{R}&\textbf{F1}&\textbf{P}&\textbf{R}&\textbf{F1}&\textbf{P}&\textbf{R}&\textbf{F1}\\
\hline
ASV&0.59&0.60&0.59&0.44&0.47&0.45&0.32&0.34&0.35\\
Bi-LSTM(W)&0.61&0.61&0.61&0.45&0.45&0.45&0.34&0.36&0.35\\
Bi-LSTM(C)&0.63&0.63&0.63&0.48&0.47&0.47&0.42&0.39&0.40\\
DSC-T(RF)&0.34&0.35&0.34&0.31&0.32&0.31&0.24&0.23&0.23\\
MNB-H&0.45&0.49&0.46&0.42&0.43&0.42&0.38&0.36&0.37\\\hline
SSC(Eng-Eng)&\textbf{0.74}&\textbf{0.73}&\textbf{0.73}&\textbf{0.62}&\textbf{0.60}&\textbf{0.61}&\textbf{0.49}&\textbf{0.54}&\textbf{0.51}\\
SSC(Spa-Spa)&0.71&0.72&0.71&0.58&0.59&0.58&0.42&0.42&0.42\\
SSC(Hin-Hin)&0.54&0.58&0.56&0.45&0.47&0.46&0.39&0.33&0.35\\
SSC(Tel-Tel)&0.47&0.49&0.48&0.39&0.41&0.40&0.31&0.35&0.33\\
SSC(Hin-Eng)&\textbf{0.68}&\textbf{0.70}&\textbf{0.69}&\textbf{0.52}&\textbf{0.56}&\textbf{0.54}&\textbf{0.46}&\textbf{0.43}&\textbf{0.44}\\
SSC(Tel-Eng)&\textbf{0.63}&\textbf{0.66}&\textbf{0.64}&\textbf{0.49}&\textbf{0.47}&\textbf{0.48}&\textbf{0.41}&\textbf{0.44}&\textbf{0.42}\\
\end{tabular}
\caption{Comparison between different language pairs of our model and previous methodologies for emoji prediction experiment. 5,10 and 18 are the number of most frequent emojis considered in that experiment. P,R,F1 are the Precision, Recall and F-scores respectively.}
\label{tab:emoji prediction}
\end{table*}
\subsection{Experimental Setup}
We validate our models on the tasks of sentiment analysis and multilingual emoji prediction. Both of these setups are evaluated seperately.
\subsubsection{Sentiment Analysis}
The experiment is a text classification task. We take the English and Hindi sentiment datasets (Eng-Hin) and align each Hindi sentence with English sentences of the same sentiment (positive samples) and label them 1. Similarly, we also randomly sample the same number of English tweets with different sentiment (negative samples) for each Hindi Tweet and label them -1. Similarly, we repeat the experiment for all the different pairs of datasets possible.

For the comparative study, we train the baselines defined in section \ref{sec:baselines} on the same datasets. In case of resource-rich languages (English and Spanish), the baselines ASV and MV-RNN are trained on their respective sentiment datasets. We train the baseline MNB-H on Hindi sentiment dataset and baseline DSC-T on Telugu sentiment dataset. The baselines of English compare to the (Eng-Eng) pair's performance. For each of the other languages \textit{lang}, we compare their baselines with the performance (Eng-\textit{lang}) and (\textit{lang}-\textit{lang}) pair. Table \ref{tab:sentiment analysis} demonstrates the results of sentiment analysis experiments.
\subsubsection{Emoji Prediction}
We perform contrastive learning on our model using data made by aligning each English tweet with a set of positive Hindi tweet samples (with the same emoji) with label 1 and a set of negative Hindi tweet samples (with different emoji) of the same size with label -1. We conduct this experiment thrice taking 5, 10 most frequent emojis and all the emojis (18 classes). Similarly, we repeat the experiment for all the different pairs of datasets possible.

For an appropriate comparative study, we also train the baselines defined in section \ref{sec:baselines} on the twitter-emoji datasets. In case of resource-rich languages (English and Spanish), the baseline Bi-LSTM model is trained on the respective twitter-emoji datasets. We train the baseline MNB-H on Hindi sentiment dataset and baseline DSC-T on Telugu twitter-emoji dataset. The baselines of English compare to the (Eng-Eng) pair's performance. For each of the other languages \textit{lang}, we compare their baselines with the performance (Eng-\textit{lang}) and (\textit{lang}-\textit{lang}) pair. Table \ref{tab:emoji prediction} demonstrates the results of the emoji prediction experiments.
\section{Analysis of Results}
\label{sec:results}
\subsection{Qualitative Analysis}
As we observe from the results (given in tables \ref{tab:sentiment analysis} and \ref{tab:emoji prediction}) of same language pairs (Eng-Eng,Spa-Spa,Hin-Hin,Tel-Tel), our model outperforms its counterparts significantly on both the tasks despite the amount of data being same. The improvement in performance is also observed for both the tasks. The reason is that our model is centered around the similarity metric. Bi-LSTMs learn representations of the sentences according to the similarity metric. This helps the model in capturing task-specific features along with necessary semantic features. The shared parameters of the network enable the sentences' projection to the same problem space based on the similarity metric.  
\subsection{Quantitative Analysis}
From the overall results (given in tables \ref{tab:sentiment analysis} and \ref{tab:emoji prediction}), we observe that the model's performance in both the tasks is directly proportional to the amount of data available in the language. Also, we observe that in the cases where we pair resource-poor languages with relatively resource-rich languages (Eng-Hin, Eng-Tel), there is a significant improvement in performance compared to their monolingual performance. The Bi-LSTMs and their shared parameters allow the system to project sentences into the same problem space. This allows the problem space to be language-agnostic. Hence, the abundant resources of other languages are able to promote significant performance improvements in resource-poor languages.
\section{Conclusion}
\label{sec:conclusion}
In this paper, we introduce a solution to leverage resource-rich languages and enhance text classification in resource-poor languages. For this, we proposed an architecture that solves the problem by projecting language-pairs into the same problem space. The model employs twin Bi-LSTM networks with shared parameters to capture a task-specific representation of the sentences. These representations are utilized in conjunction with a similarity metric to group sentences with similar classes together.
The qualitative analysis showed that our model outperforms the state-of-the-art methodologies with training on same language pairs. The quantitative analysis presented a significant enhancement in the model's prediction by training the resource-poor language in conjunction with resource-rich language.
\bibliographystyle{named}
\bibliography{ijcai18}

\end{document}